# Multi-vision-based Picking Point Localisation of Target Fruit for Harvesting Robots

C. Beldek
School of Mechanical, Materials,
Mechatronic, and Biomedical
Engineering
University of Wollongong
Wollongong, NSW, Australia
cbeldek@uow.edu.au

A. Dunn
School of Mechanical, Materials,
Mechatronic, and Biomedical
Engineering
University of Wollongong
Wollongong, NSW, Australia

J. Cunningham
School of Mechanical, Materials,
Mechatronic, and Biomedical
Engineering
University of Wollongong
Wollongong, NSW, Australia

E. Sariyildiz
School of Mechanical, Materials,
Mechatronic, and Biomedical
Engineering
University of Wollongong
Wollongong, NSW, Australia

S. L. Phung
School of Electrical, Computer and
Telecommunications
Engineering
University of Wollongong
Wollongong, NSW, Australia

G. Alici
School of Mechanical, Materials,
Mechatronic, and Biomedical
Engineering
University of Wollongong
Wollongong, NSW, Australia

*Abstract*—This paper presents multi-vision-based localisation strategies for harvesting robots. Identifying picking points accurately is essential for robotic harvesting because insecure grasping can lead to economic loss through fruit damage and dropping. In this study, two multi-vision-based localisation methods, namely the analytical approach and model-based algorithms, were employed. The actual geometric centre points of fruits were collected using a motion capture system (mocap), and two different surface points $C_{fix}$ and $C_{eih}$ were extracted using two Red-Green-Blue-Depth (RGB-D) cameras. First, the picking points of the target fruit were detected using analytical methods. Second, various primary and ensemble learning methods were employed to predict the geometric centre of target fruits by taking surface points as input. Adaboost regression, the most successful model-based localisation algorithm, achieved 88.8% harvesting accuracy with a Mean Euclidean Distance (MED) of 4.40 mm, while the analytical approach reached 81.4% picking success with a MED of 14.25 mm, both demonstrating better performance than the single-camera, which had a picking success rate of 77.7% with a MED of 24.02 mm. To evaluate the effect of picking point accuracy in collecting fruits, a series of robotic harvesting experiments were performed utilising a collaborative robot (cobot). It is shown that multi-vision systems can improve picking point localisation, resulting in higher success rates of picking in robotic harvesting.

## I. Introduction

With rising challenges such as increasing the world population, aging society, labour shortage and labour cost, agricultural robots have gained importance in modern farming [1-3]. Robotic picking is an innovative technology that allows harvesting operations to be executed gently, accurately, quickly and without human intervention [4]. Open agricultural settings are typically unpredictable and suffer from complex environmental disturbances, hindering accurate localisation and successful picking [5, 6]. To overcome these challenges and execute harvesting tasks selectively and properly, robotic pickers are equipped with various sensors and hardware [7, 8].

Advancements in camera and optical lens technology have made vision-based perception strategies more cost-effective and practical for harvesting operations [9, 10]. The main goal of employing cameras in harvesting operations is to estimate the position and orientation of target fruits. In literature, numerous types of cameras have been used for this purpose. For example, Chen et al. proposed a Convolutional Neural Networks (CNNs)-based multi-target positioning approach using a monocular Red-Green-Blue (RGB) camera [11]. However, monocular RGB cameras cannot measure distance during picking tasks without including additional methods such as probabilistic graph models and deep learning models [12]. Although depth information can be obtained using a binocular stereo configuration, it is more computationally complex and needs at least two sensors [13]. Hence, RGB-D cameras have become the mainstream machine vision hardware in agricultural robotic harvesting operations with their measuring depth capability, high applicability and affordability [9, 14, 15]. The integrated use of RGB-D cameras with machine learning and deep learning models has demonstrated effective performance in crop detection. For instance, Rathore et al. proposed a novel two-stage deep learning-based detection and classification algorithm that uses an RGB-D camera to classify apples based on their occlusion level [16]. Nevertheless, detecting and localising a target fruit remains a bottleneck in robotic harvesting although there are advancements in vision-based sensing tools and software. [17]. This is primarily due to environmental effects that lead to disturbances, e.g., illumination variability, occlusion, overlapping and background disturbances [18-21].

Fruit detection and identifying the appropriate picking point are critical in harvesting, as insecure grasping can lead to fruit damage and dropping while carrying fruit [22-24]. For example, the geometric centre (G) can serve as the picking point for spherical fruits assuming homogeneity [25]. However, the centre points of fruits were mostly extracted and projected from 2D image data. The identified points are surface points on the fruit and do not represent the actual picking points needed for harvesting operations. Studies using 2D image data include Abeyrathna et al. [26], who achieved a relatively high root mean square error (RMSE) of a minimum of 1.54 cm in their study. Additionally, Hu et al. [27] applied an offset equal to the radius of the fruit to the depth-axis along the surface after detecting the picking point on the surface. This can lead to insecure grasping during harvesting operations because each fruit possesses a unique shape and growth form in the orchards [28]. Therefore, identifying picking points should incorporate more advanced methods rather than relying on 2D approaches. Integrating gripper design with a wider grasping range to picking point identification can present a solution to tolerate localisation errors for improving picking applications [29]. Nevertheless, this can result in grasping fruit with other tree parts [30]. As another solution, multi-sensor-based approaches have been proposed to improve localisation performance in robotic harvesting e.g., lidar-camera fusion can perform precise and robust depth-sensing in real orchards [31] and multi-camera-based sensor fusion methods can improve the localisation performance of the harvesting system [32]. Despite the effectiveness of multi-sensor-based localisation methods, further efforts are required to enhance their accuracy in harvesting practices.

In this study, two multi-camera-based localisation methods are proposed to improve the success rate in robotic harvesting. Actual picking points were obtained employing the mocap system, and surface points were collected employing RGB-D cameras. Utilising surface points, an analytical multi-camera picking point identification method has been proposed. Furthermore, machine learning models as both primary and ensemble learning techniques were utilised to predict picking point coordinates. All prediction methods were tested and verified with robotic hardware in real harvesting operations. The results emphasised that multi-vision can enhance harvesting accuracy while enabling successful fruit localisation.

The remainder of the paper is organised as follows. Section II explains the robotic harvesting system in detail. Section III describes the comprehensive methodology of picking point identification methods including analytical methods and model-based algorithms. Section IV presents the conducted experiments with data collection, results and analyses. Section V elucidates the harvesting application, and finally, Section VI provides the conclusion of the research study.

## II. ROBOTIC HARVESTING SYSTEM

### A. Harvesting System

  a) *Robotic Hardware:* The robotic hardware in this study consists of 7 Degree of Freedom (DoF) Emika Franka Research-3 cobot and two-finger parallel gripper Franka Hand. Additionally, two Intel RealSense D415 RGB-D cameras were employed as a sensor hardware with the robotic system.
  b) *Perception Network:* To achieve accurate detection performance, a YOLOv8-n pre-trained on the COCO (Common Objects in Context) dataset was used in this study. The model was run on a Dell OptiPlex 7040 with an Intel® Core™ i7-6700 CPU @ 3.40 GHz, 8 GB RAM, and an AMD® Oland XT (Radeon HD 8670 / R7 250/350) GPU to detect oranges on the tree.

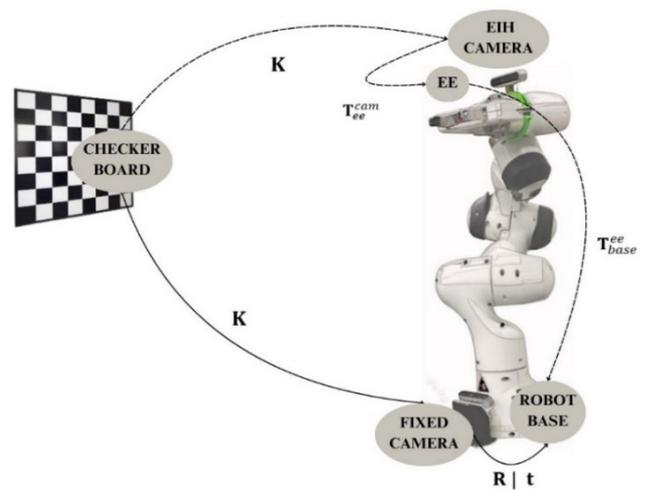

Fig. 1. Multi-vision localisation

### B. Multi-vision Localisation

Two different RGB-D cameras were implemented in this study. In contrast, one of the cameras was attached to the robotic arm in an eye-in-hand configuration, the other RGB-D camera was fixed on the mobile platform.

The image projection model, which enables the transformation of pixel points to the world coordinates, is expressed in Eq. (1).

$$\begin{bmatrix} u \\ v \\ w \end{bmatrix} = \begin{bmatrix} \mathbf{K} & \mathbf{0} \\ \mathbf{0} & 1 \end{bmatrix} \begin{bmatrix} \mathbf{R} & \mathbf{t} \\ \mathbf{0} & 1 \end{bmatrix} \begin{bmatrix} X \\ Y \\ Z \\ 1 \end{bmatrix} \quad (1)$$

where $u$ and $v$ denote the horizontal and vertical pixel coordinates respectively; $w$ is the homogeneous scale factor in the image frame, which is directly proportional to the depth in the camera coordinate system; $X, Y$ and $Z$ represent the corresponding coordinates in the world coordinate frame. The intrinsic camera matrix is defined as $\mathbf{K} = \begin{bmatrix} f_x & 0 & c_x \\ 0 & f_y & c_y \\ 0 & 0 & 1 \end{bmatrix}$, where $f_x$ and $f_y$ are the focal lengths along the horizontal and vertical axes, and $c_x$ and $c_y$ are the coordinates of the principal point in pixels. The transformation matrix $\mathbf{R} \mid \mathbf{t} = \begin{bmatrix} r_{11} & r_{12} & r_{13} & t_1 \\ r_{21} & r_{22} & r_{23} & t_2 \\ r_{31} & r_{32} & r_{33} & t_3 \end{bmatrix}$ combines the rotation matrix $\mathbf{R}$ and translation vector $\mathbf{t}$. In this matrix, $r_{ij}$ corresponds to the element in the $i$-th row and $j$-th column of the rotation matrix, while $t_i$ indicates the $i$-th component of the translation vector. To estimate $\mathbf{K}$ and $\mathbf{R} \mid \mathbf{t}$, intrinsic and extrinsic camera calibration were conducted for both cameras.

*1) Intrinsic Camera Calibration:* As both of the cameras are the same model and intrinsic camera calibration is unique to the manufacturer and camera model, intrinsic camera calibration was conducted only once to extract the intrinsic parameters ($f_x, f_y, c_x$ and $c_y$).

*2) Extrinsic Camera Calibration:* Extrinsic parameters, defined as rotation and translation, describe the orientation

and position of the camera, respectively. As two different camera configurations were employed in this study, two distinct extrinsic camera matrices were derived, as shown in Fig. 1.

- a) *Fixed camera:* Due to the fixed camera position and its proximity to the robot base origin, $R \mid t$ is extracted manually. The fixed camera is mounted on a holder with a rotation of $R_{fix}(-60°)$ indicating a $-60°$ rotation around the $Y$-axis w.r.t. the robot base coordinate frame. Additionally, $[t_X \quad t_Y \quad t_Z]^T$ were added to rotated coordinates to determine the positional information, where $[t_X \quad t_Y \quad t_Z]^T$ is the transpose of the translation offset vector for the $X, Y$ and $Z$ axes respectively in mm.
- b) *Eye-in-hand camera:* The RGB-D camera is attached between the flange joint and the 7th joint of the robotic arm. This setup means that the camera moves during the harvesting operation, and the extrinsic camera matrix changes for each camera pose. $T_{ee}^{cam}$ (camera-to-end-effector homogeneous transformation matrix) was extracted using a hand-eye calibration strategy with 200 iterations by the VISP library [33]. Although the same checkerboard was used to define the world coordinates, the corresponding camera positions were also determined by the end-effector coordinate frame for each pose. Once the camera-to-end-effector homogeneous transformation matrix is obtained, the $T_{base}^{cam}$ (camera-to-robot base homogeneous transformation matrix) can be derived using Emika Franka's kinematic chain as shown in Eq. (2):

$$\mathbf{T}_{base}^{cam} = \mathbf{T}_{base}^1 \mathbf{T}_1^2 \mathbf{T}_2^3 \mathbf{T}_3^4 \mathbf{T}_4^5 \mathbf{T}_5^6 \mathbf{T}_6^7 \mathbf{T}_7^{ee} \mathbf{T}_{ee}^{cam} \quad (2)$$

where $\mathbf{T}_i^j$ is the homogeneous transformation matrix that transforms the position and orientation of the $j$-th joint of the robotic arm w.r.t $i$-th joint.

### III. PICKING POINT DETECTION METHODS

#### A. Analytical Methods

*1) Single-camera*

In the single-camera-based detection methods, the centre point of the fruit's bounding box was extracted by the YOLOv8 perception network. Using the approximate radius of the fruit, an offset is added to a particular axis along the surface as recommended in [27]. In this study, an eye-in-hand camera was used as a single-camera configuration and the picking point coordinates were obtained using Eq. (2).

*2) Multi-camera*

Once collecting $C_{fix}$ and $C_{eih}$, a multi-camera-based localisation method was employed to calculate the picking points of the fruit analytically as illustrated in Fig. 2

In this figure, $d$ is the line segment passing through the points $C_{fix}$ and $C_{eih}$; $M$ represents the midpoint of the $d$ and it can be calculated as $M = \frac{C_{fix} + C_{eih}}{2}$; r represents the radius of the target fruit which is 35 mm for an artificial orange for this setup; and $\mathbf{R}_{fix}$ and $\mathbf{R}_{eih}$ represent the rotation matrix of fixed and eye-in-hand cameras, respectively. Once collecting $C_{fix}$ and $C_{eih}$, predicted picking point coordinates can be derived as shown in Eq. (3):

$$G = M + r\left(\frac{\mathbf{dir_{fix}} + \mathbf{dir_{eih}}}{2}\right) \quad (3)$$

where $\mathbf{dir_{fix}}$ and $\mathbf{dir_{eih}}$ represent rotation vectors that were obtained from $\mathbf{R_{fix}}$ and $\mathbf{R_{eih}}$ using Rodrigues' rotation formula for fixed and eye-in-hand cameras, respectively.

#### B. Model-based Methods

*1) Primary Learners*

Various machine learning techniques were implemented to estimate G point coordinates by detecting $C_{fix}$ and $C_{eih}$. Linear regression (LR) served as the baseline method to capture the linear relationships of surface points and picking points. In Support Vector Regression (SVR) with a linear kernel and regularisation parameter, C=10.0 was employed to model both linear and slight non-linear relationships, handling minor deviations. DT (Decision Tree), with standard scikit-learn regression settings, predicted the picking point coordinates by splitting data into decision rules for interpretability. Finally, a Multi-Layer Perception (MLP) regressor that has a single hidden layer with 100 neurons for each model was utilised to estimate picking point coordinates. The ReLU activation function and Adam optimiser with 200 maximum iteration numbers were used in MLP architecture.

*2) Ensemble Learners*

Ensemble learning techniques were employed to improve the estimation accuracy of the picking point by leveraging multiple base learners. These methods combine predictions from different models to produce more reliable results. Various key regression models were applied:

- a) *Adaboost:* Three models using LR as the base estimator were trained, each with 100 estimators and a learning rate of 0.1. AdaBoost assigns higher weights to mispredicted instances, refining accuracy with each iteration. The models were evaluated on datasets on $X_1, X_2,$ and $X_3$, to analyse the effects of boosting on prediction accuracy.
- b) *Bagging:* Two models were implemented with LR and DT as base estimators, each using 100 estimators. LR reduced variance across datasets by averaging predictions, while DT captured non-

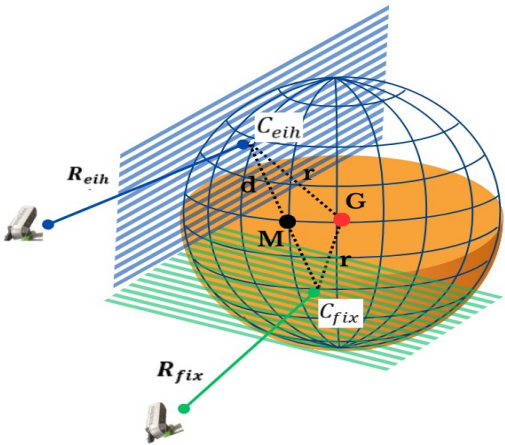

Fig. 2. Analytical multi-camera-based picking point detection method

linearities by training on data subsets, improving robustness and stability.

c) *Random forest(RF):* models that have the same structure for each axis were trained with 100 tree estimators and a learning rate of 0.1 to control the contribution of each tree.

d) *Voting Regression(VR):* This model combined predictions from multiple learners, including LR, DT, Gradient Boosting, AdaBoost, Bagging, and RF while the meta-learner ElasticNet (alpha=1.0) balanced regularisation to prevent overfitting. Three Voting models were trained on different datasets ($X_1$, $X_2$, and $X_3$) to evaluate the ensemble's ability to handle varying data complexities.

e) *Gradient boosting:* Three models that have the same structure for each axis were trained with 100 estimators and a learning rate of 0.1 to control the contribution of each tree.

f) *Stacking Regression(SR):* The prediction models consist of various base learners (DT, Gradient boosting, Adaboost, Bagging and RF), and as a meta-learner model, LR was employed for the final prediction.

## IV. EXPERIMENTS

### A. Data Collection

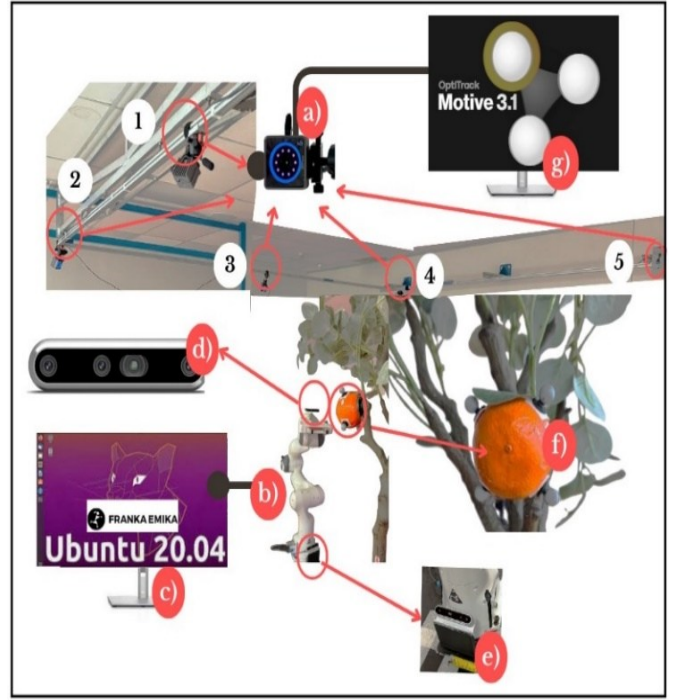

Fig. 3. Experimental setup

TABLE I
LOCALISATION ERRORS OF GEOMETRIC CENTERS FOR THE TEST POINTS

| Model | MAE x(mm) | MAE y(mm) | MAE z(mm) | MED (mm) |
|---|---|---|---|---|
| LR | 4.00 | 1.74 | 0.97 | 4.46 |
| DT | 10.31 | 16.02 | 1.56 | 19.11 |
| SVR | 4.02 | 1.70 | 1.03 | 4.48 |
| MLP | 6.65 | 4.59 | 6.29 | 10.23 |
| AdaBoost | 3.71 | 2.13 | 1.07 | 4.40 |
| Bagging (DT) | 6.22 | 7.88 | 2.36 | 10.31 |
| Bagging (LR) | 3.97 | 1.72 | 0.95 | 4.42 |
| RF | 6.21 | 9.11 | 3.03 | 11.43 |
| VR | 3.77 | 4.05 | 1.39 | 5.70 |
| Gradient boosting | 6.75 | 9.32 | 2.71 | 11.82 |
| SR | 4.68 | 1.68 | 1.92 | 5.33 |
| Analytical multi-camera | 9.87 | 5.59 | 8.63 | 14.25 |
| Eye-in-hand camera | 19.24 | 8.05 | 11.92 | 24.02 |

The mocap data comprises three types of positional information. The first set of data consists of the X, Y and Z coordinates of the G point on the fruit, captured using the mocap system. The experimental setup employs five Optitrack Prime-13 (NaturalPoint, Corvallis, OR, USA) cameras connected to a computer via a USB hub. These cameras were evenly spaced around the robotic setup and artificial tree and they are aligned to face the fruit marker configuration as can be seen in Fig.3. The mocap system was calibrated using a CW-500 Calibration Wand, CS-400 Calibration Square and Motive v3.1.1 software, achieving a mean ray error of 0.754 mm and a mean wand error of 0.199 mm. Once the mocap was calibrated, origin and fruit markers were added to the harvesting setup with some extra markers for known points in the setup to make localisation easy and accurate.

The camera data comprises the X, Y and Z coordinates of the bounding boxes' centres ($C_{fix}$ and $C_{eih}$), collected using both fixed and eye-in-hand cameras for each pose, respectively. In summary, actual picking point coordinates are obtained through the mocap, while the surface points on the fruit are detected using fixed and eye-in-hand cameras. Picking point coordinates and pairs of $C_{fix}$ and $C_{eih}$ were recorded for a total of 48 fruit poses.

The dataset consists of fixed-camera coordinates ($x_{fix}$, $y_{fix}$, $z_{fix}$), eye-in-hand camera coordinates ($x_{eih}$, $y_{eih}$, $z_{eih}$), and the target coordinates ($x_G, y_G$ and $z_G$). The data is organised into three sets: $X_1$, with $x_{fix}$ and $x_{eih}$, $X_2$ with $y_{fix}$ and $y_{eih}$, and $X_3$ with $z_{fix}$ and $z_{eih}$. The corresponding targets are $y_1$ for $x_G$, $y_2$ for $y_G$, and $y_3$ for $z_G$. The data is split into training and testing sets using an 80-20 ratio.

### B. Results

When analysing the accuracy of primary and ensemble learning models, the axis-based mean absolute errors (MAE) and MED reveal how these regressors compare to each other. The results of the picking point localisation accuracy experiments are presented in Table. I.

The evaluation of LR, DT and SVR showed distinct performance across the x, y and z axes. LR achieved the lowest MED of 4.15 mm, demonstrating the best overall accuracy with an MAE of 4.00 mm, 1.74 mm and 0.97 mm for the x, y and z axes, respectively. DT had a higher MED of 19.11 mm with larger MAE values, except for a low 1.56 mm on the z-axis. SVR achieved a MED of 4.48 mm but did not surpass LR in accuracy, with MAEs of 2.60 mm, 3.04 mm and 1.82 mm across the axes. MLP displayed balanced performance with 6.65 mm, 5.59 mm and 6.29 mm for the x, y and z axes, respectively. LR's effectiveness on this near linear dataset was clear, highlighting each model's strengths and limitations.

For ensemble learning, Adaboost had a MED of 4.40 mm, with MAE values of 3.71 mm on the x-axis, 2.13 mm on the y-axis, and 1.07 mm on the z-axis. Adaboost outperformed DT, handling nonlinear data more effectively and achieving the lowest error among all learners. DT-based Bagging had one of the highest MEDs at 10.31 mm, with notable errors on the x and y axes, though lower than the DT model. LR-based Bagging achieved a MED of 4.42 mm, with the lowest z-axis MAE of 0.95 mm, indicating strong performance on this dataset.

VR model showed balanced performance with a MED of 3.95 mm and relatively low MAEs across all axes, leveraging strengths from LR, DT and SVR. SR had a MED of 5.33 mm, with MAEs of 4.68 mm, 1.68 mm and 1.92 mm across the x, y and z axes, respectively.

The least accurate ensemble models were RF and Gradient Boosting, with MED values of 11.43 mm and 11.82 mm, respectively.

The success of traditional ML models, both primary and ensemble, is closely linked to dataset structure. In this study, the dataset assumed the target fruit as a 35 mm-radius sphere, resulting in a high linearity. Models that are adept at identifying linear patterns performed best. For instance, LR outperformed the non-linear DT by over four times. Similarly, LR-Bagging had narrower error ranges than DT-Bagging when modelling this near-linear data. SR, one of the most accurate models, effectively combined base learners with an LR meta-learner. LR's role was to optimally aggregate base learner predictions. In contrast, models that use tree structures (DT, RF and Gradient Boosting) were designed to capture non-linear patterns and showed the least accuracy. Unlike non-linear datasets, linear relationships can be analytically defined. However, the analytical multi-camera strategy performed poorly, with a MED of 14.28 mm and MAEs of 9.87 mm, 5.59 mm, and 8.63 mm for the x, y and z axes, respectively. The error in the analytical multi-camera strategy stems from experimental factors, especially the spherical fruit shape assumption.

## V. Application: Orange Harvesting

Model-based and analytical multi-camera localisation strategies can be employed in real harvesting practices as a picking point extraction method via surface points. The findings derived from the testing dataset confirm that each error range is within acceptable limits for accurately executing picking operations [34]. To confirm the proposed picking point detection methods, an offline robotic harvesting application has been performed, illustrated in Fig. 4.

In this application, the eye-in-hand camera, Adaboot regression which is the best model-based method and the analytical multi-camera localisation strategy have been tested in controlled conditions with an artificial fruit-tree setup. Each experiment had been completed with 27 random fruit positions, and picking attempts had been executed triple for each fruit position. In eye-in-hand camera-guided harvesting operations 35 mm z-axis offset was added to cartesian coordinates as recommended in [27]. The picking successes were 77.7, 81.4 and 88.8 % for the eye-in-hand camera, the analytical multi-camera strategy and Adaboost regressor, respectively. Thereby, the proposed picking point detection methods have been practically verified by the orange harvesting application.

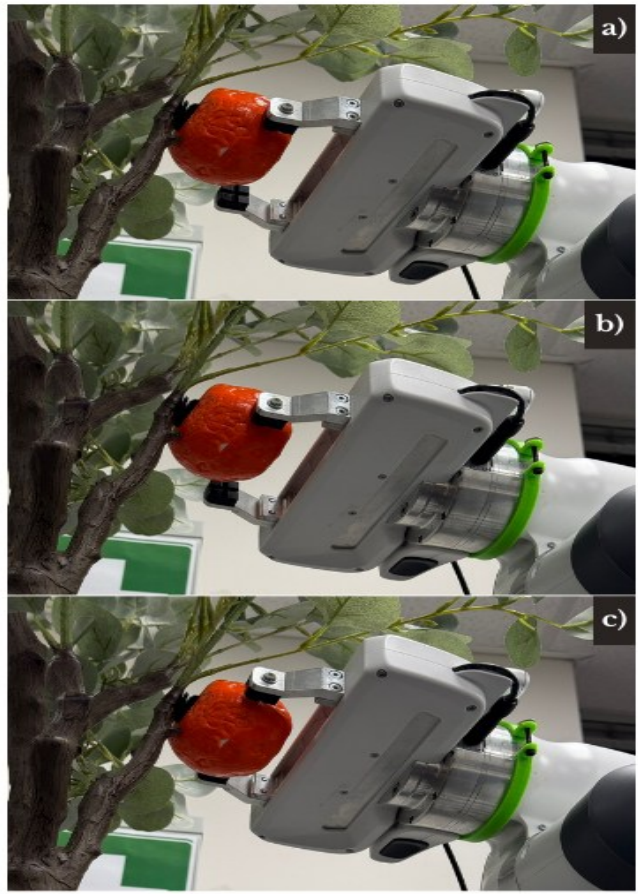

Fig. 4. Orange harvesting application via eye-in-hand camera localisation (a), the analytical method (b) and Adaboost regression (c) picking point detection

According to the results, multi-camera-based localisation strategies can define picking points more accurately which is significant to estimate grasping position during harvesting. Therefore, multi-vision-based localisation strategies can improve picking accuracy in robotic harvesting practices. An example of successful picking attempts via eye-in-hand camera localisation (a), the analytical multi-camera method (b) and Adaboost regression (c) picking point detection in the harvesting application have been shown in Fig. 4. As shown in this figure, positioning the picking point further from the target's geometric centre toward the surface increases the risk of insecure grasping. In terms of insecure grasping risk, single-camera ranks the highest, followed by the analytical multi-camera method and then the model-based method. To this end, the findings obtained from both the successful and unsuccessful harvesting attempts and the test point localisation experiments are in line.

## VI. Conclusion

In this study, we have demonstrated multi-vision-based positioning strategies using analytical approaches and model-based methods for spherical fruits. Positioning target fruits is challenging under the environmental disturbances in agricultural fields. A series of experimental tests have been presented in this paper to verify the localisation performance of multi-vision-based strategies. While the best model, AdaBoost regression, has a MED of 4.40 mm, the analytical multi-camera strategy has a MED of 14.25 mm. Furthermore, multi-vision-based picking point detection methods demonstrated higher accuracy in the harvesting application compared to the single-camera-based method. The analytical multi-camera method improved picking accuracy by 3.7%, while AdaBoost regression enhanced it by 11.1%. Further studies should be carried out on the accurate localisation of crops with unique shapes, partially hidden and under the effect of various disturbances. Further studies should be carried out on the accurate localisation of crops with unique shapes, partially hidden and under the effect of various disturbances to provide a better understanding.